\documentclass[10pt,twocolumn,letterpaper]{article}

\usepackage{cvpr}              

\definecolor{cvprblue}{rgb}{0.21,0.49,0.74}
\usepackage[pagebackref,breaklinks,colorlinks,allcolors=cvprblue]{hyperref}

\usepackage{algpseudocode}
\usepackage[ruled,vlined]{algorithm2e}
\usepackage{placeins}

\def\paperID{7}
\def\confName{CVPR}
\def\confYear{2026}

\title{Adaptive Data Dropout: Towards Self-Regulated Learning in Deep Neural Networks}

\author{
Amar Gahir$^{*}$ \hspace{1em}
Varshil Patel$^{*}$ \hspace{1em}
Shreyank N Gowda\\
School of Computer Science, University of Nottingham, Nottingham, United Kingdom\\
{\tt\small shreyank.narayanagowda@nottingham.ac.uk}
}

\begin{document}
\maketitle
\begin{abstract}
Deep neural networks are typically trained by uniformly sampling large datasets across epochs, despite evidence that not all samples contribute equally throughout learning. Recent work shows that progressively reducing the amount of training data can improve efficiency and generalization, but existing methods rely on fixed schedules that do not adapt during training. In this work, we propose Adaptive Data Dropout, a simple framework that dynamically adjusts the subset of training data based on performance feedback. Inspired by self-regulated learning, our approach treats data selection as an adaptive process, increasing or decreasing data exposure in response to changes in training accuracy. We introduce a lightweight stochastic update mechanism that modulates the dropout schedule online, allowing the model to balance exploration and consolidation over time. Experiments on standard image classification benchmarks show that our method reduces effective training steps while maintaining competitive accuracy compared to static data dropout strategies. These results highlight adaptive data selection as a promising direction for efficient and robust training. Code will be released.
\end{abstract}    
\section{Introduction}
\footnote{$^{*}$Equal contribution. Work done during the DeepMind Research Ready Internship.}
Advances in deep learning have consistently come at the cost of substantial computational resources. Each year, the computational cost of the most notable models multiplies by over 4 times~\cite{epoch2023aitrends}. This trend raises significant challenges for real-world deployment, scalability, and environmental sustainability~\cite{gowda2023watt}. While a large body of work has focused on improving model efficiency through architectural design, such as MobileNet~\cite{howard2017mobilenets, sandler2018mobilenetv2}, EfficientNet~\cite{tan2019efficientnet}, and EfficientFormer~\cite{li2022efficientformer}, as well as techniques like pruning~\cite{han2015learning}, quantization~\cite{jacob2018quantization}, and knowledge distillation~\cite{hinton2015distilling}, relatively less attention has been paid to improving the efficiency of the training process itself.

Standard training pipelines treat all samples equally across all epochs, implicitly assuming that every data point contributes uniformly throughout learning. However, prior work has shown that this assumption is flawed, as some samples provide diminishing returns once they are learned, while others remain informative for longer~\cite{katharopoulos2018not}. Inspired by this observation, recent work introduced Progressive Data Dropout (PDD)~\cite{shriramprogressive}, a simple yet effective approach that progressively reduces the amount of training data across epochs. By focusing on informative samples early and revisiting the full dataset later, PDD achieves substantial reductions in effective training cost while maintaining or improving generalization.

Despite its effectiveness, PDD relies on pre-defined or pre-computed data schedules that remain fixed throughout training. These schedules, whether difficulty-based or stochastic, do not adapt to the evolving state of the model. As a result, they may either drop data too aggressively, leading to underfitting, or retain redundant samples longer than necessary, reducing potential efficiency gains. More fundamentally, fixed schedules fail to capture a key aspect of human learning: the ability to dynamically adjust study strategies based on ongoing performance.

\begin{figure}
    \centering
    \includegraphics[width=0.99\linewidth]{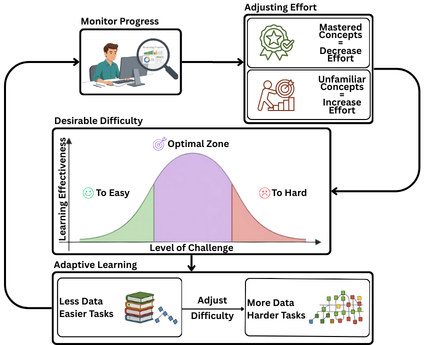}
    \caption{Adaptive learning balances challenge by dynamically adjusting effort based on performance feedback.}
    \label{fig:teaser}
\end{figure}

Human learning is inherently adaptive. Learners monitor their progress and regulate effort accordingly, allocating more attention when performance deteriorates and reducing effort when concepts are well understood. This process, often described as self-regulated learning~\cite{zimmerman2002becoming}, is closely linked to the notion of desirable difficulties~\cite{bjork2011making}, where learning is most effective when tasks are neither too easy nor too difficult. The optimal level of challenge is not static but evolves as the learner improves, requiring continuous adjustment of learning conditions~\cite{guadagnoli2004challenge}. These insights suggest that efficient learning systems should not rely on fixed curricula, but instead adapt the level of difficulty and data exposure over time. Figure~\ref{fig:teaser} illustrates this visually.

In this work, we propose \textit{Adaptive Data Dropout}, a simple and general framework that extends PDD by introducing feedback-driven data scheduling. Rather than following a fixed dropout schedule, our method dynamically adjusts the subset of training data at each epoch based on observed learning progress. Concretely, we use changes in training performance as a signal to modulate the degree of data dropout, allowing the model to automatically increase or decrease data exposure as needed. This results in a training process that maintains an appropriate level of difficulty throughout learning, balancing efficiency and generalization without requiring pre-defined schedules.

Our approach retains the simplicity and compatibility of PDD, requiring no changes to model architectures or optimization procedures, while introducing adaptivity into the training loop. We demonstrate that adaptive data dropout consistently reduces effective training cost while maintaining competitive performance across architectures. These results highlight the importance of feedback-driven data selection and suggest a promising direction toward more efficient and human-like learning dynamics in deep neural networks.

\section{Related Work}

A broad range of research efforts aim to improve training efficiency and generalization in deep learning. These include advances in model design, data selection, and training strategies. We situate our work within these directions, with particular focus on methods that adapt data usage during training.

\paragraph{Model Efficiency and Training Cost.}
A substantial body of work has focused on reducing the computational footprint of neural networks, primarily at inference time. Techniques such as pruning~\cite{han2015learning}, quantization~\cite{jacob2018quantization}, and knowledge distillation~\cite{hinton2015distilling} aim to compress models while preserving performance. Similarly, architectures such as MobileNet~\cite{howard2017mobilenets}, EfficientNet~\cite{tan2019efficientnet}, and EfficientFormer~\cite{li2022efficientformer} are designed to minimize parameters and FLOPs. In contrast, our work targets the training process itself, aiming to reduce the amount of data used during optimization while maintaining or improving performance, without modifying the underlying model.

\paragraph{Curriculum Learning and Adaptive Sampling.}
Curriculum learning~\cite{bengio2009curriculum} and hard example mining~\cite{shrivastava2016training} organize training samples based on difficulty, typically presenting easier or more informative examples at specific stages of learning. Extensions of this idea explore adaptive sampling strategies that prioritize samples based on loss, gradient magnitude, or uncertainty~\cite{katharopoulos2018not}. While these approaches adjust the importance or order of samples, they generally assume a predefined or heuristic schedule. In contrast, our method treats data selection as a feedback-driven process, dynamically adjusting data exposure based on observed learning progress rather than following a fixed curriculum.

\paragraph{Data Subset Selection and Pruning.}
Another line of work investigates training on reduced subsets of data to improve efficiency. Methods such as Data Diet~\cite{tonevaempirical} identify and remove less informative samples based on forgetting statistics or gradient-based metrics. Dataset pruning approaches~\cite{yang2022dataset} aim to construct compact subsets that preserve the performance of the full dataset, often requiring additional optimization or proxy models~\cite{colemanselection}. Influence-based methods~\cite{koh2017understanding, feldman2020neural} estimate the contribution of individual samples, while coreset selection techniques~\cite{mirzasoleiman2020coresets, mindermann2022prioritized} identify representative subsets through repeated evaluation. These approaches typically involve pre-processing stages or additional computational overhead. In contrast, our method operates entirely within the training loop and adapts data usage online, without requiring auxiliary models or explicit sample scoring.

\paragraph{Active Learning.}
Active learning~\cite{settles2009active} seeks to reduce labeling and training costs by selecting the most informative samples from an unlabeled pool. Common strategies include uncertainty sampling~\cite{nguyen2022measure}, margin-based selection~\cite{balcan2007margin}, and query-by-committee~\cite{kee2018query}. These methods rely on iterative querying and assume access to large unlabeled datasets. Our setting is fundamentally different, as we operate in a fully supervised regime and focus on dynamically adjusting the use of already-labeled data during training, rather than acquiring new data.

\paragraph{Data Dropout and Progressive Training.}
Dropout~\cite{srivastava2014dropout} and its variants, such as DropConnect~\cite{wan2013regularization} and stochastic depth~\cite{huang2016deep}, improve generalization by introducing stochasticity at the level of model activations. More recent work has explored analogous ideas at the data level. Approaches such as data dropout~\cite{wang2018data} and Dynamic Training Data Dropout~\cite{zhong2021dynamic} selectively remove samples during training based on heuristics or noise estimation. Progressive Data Dropout (PDD)~\cite{shriramprogressive} extends this idea by gradually reducing the training set across epochs using predefined schedules, achieving significant efficiency gains.

Our work builds directly on this line of research, but departs from prior methods in a key aspect: rather than relying on fixed or precomputed schedules, we introduce an adaptive mechanism that adjusts data dropout in response to training dynamics. This enables the model to regulate its own data exposure over time, maintaining an appropriate level of difficulty throughout training.

\paragraph{Learning with Adaptive Difficulty.}
Our work is also related to cognitive and educational theories that emphasize adaptive learning. Self-regulated learning~\cite{zimmerman2002becoming} describes how learners monitor performance and adjust their strategies accordingly. The concept of desirable difficulties~\cite{bjork2011making} suggests that learning is most effective when tasks are neither too easy nor too difficult, while the challenge point framework~\cite{guadagnoli2004challenge} formalizes the idea that optimal difficulty evolves with expertise. These perspectives highlight the importance of dynamically adjusting learning conditions, a principle that we operationalize in the context of deep neural network training through adaptive data dropout.
\section{Method}
We introduce \textit{Adaptive Data Dropout}, a simple yet effective framework that extends Progressive Data Dropout (PDD)~\cite{shriramprogressive} by dynamically adjusting data usage during training. Unlike prior approaches that rely on fixed or precomputed schedules, our method adapts the amount of training data at each epoch based on performance feedback. This enables the model to regulate its own learning process, maintaining an appropriate level of difficulty over time.

\subsection{Overview}

Let $\mathcal{D}_0 = \{(x_i, y_i)\}_{i=1}^{N}$ denote the full training dataset. At each epoch $t$, we construct a subset $\mathcal{D}_t \subseteq \mathcal{D}_0$ used for backpropagation. As in PDD, the final epoch uses the full dataset, $\mathcal{D}_T = \mathcal{D}_0$, serving as a revision phase.

In contrast to PDD, where $|\mathcal{D}_t|$ is determined by a fixed schedule, we define data usage as an adaptive process. At each epoch, the model observes a feedback signal based on training performance and updates the subset size accordingly. This allows the training procedure to dynamically increase or decrease data exposure in response to learning progress.

We define the feedback signal as:
\[
\Delta_t = \mathcal{A}_t - \mathcal{A}_{t-1},
\]
where $\mathcal{A}_t$ denotes the training accuracy at epoch $t$. This signal reflects whether the model is improving or deteriorating, and is used to guide data selection. An overview can be seen in Figure~\ref{fig:overview}.

\begin{figure}
    \centering
    \includegraphics[width=0.99\linewidth]{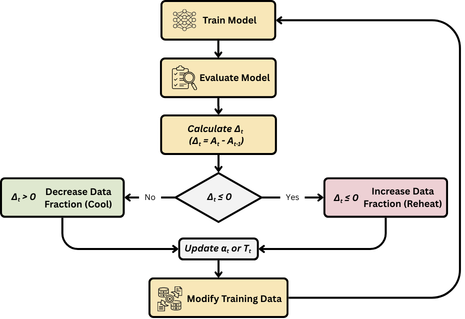}
    \caption{Adaptive data dropout dynamically adjusts training data based on changes in model performance, increasing or decreasing data exposure to maintain optimal learning difficulty.}
    \label{fig:overview}
\end{figure}

\subsection{Variant 1: Adaptive Decay (Adaptive-$\alpha$)}

Our first variant builds on scalar-based dropout by introducing an adaptive decay parameter. Instead of using a fixed decay rate, we dynamically update $\alpha_t$ based on training feedback.

At each epoch, the subset size is given by:
\[
|\mathcal{D}_t| = f(t; \alpha_t) \cdot N,
\]
where $f$ is a monotonically decreasing function (e.g., exponential, logarithmic, or inverse linear), and $\alpha_t$ controls the rate of decay.

We update $\alpha_t$ using a stochastic acceptance mechanism:
\begin{itemize}
    \item If $\Delta_t > 0$, we accept a more aggressive decay (increase $\alpha_t$), reducing the dataset size.
    \item If $\Delta_t \leq 0$, we probabilistically accept or reject the update. On rejection, we decrease $\alpha_t$, increasing data exposure in subsequent epochs.
\end{itemize}

This process allows the model to explore different dropout rates while avoiding overly aggressive data reduction. Intuitively, when learning is progressing well, the model reduces redundant data, whereas when performance stagnates, it revisits more data.

\subsection{Variant 2: Adaptive Keep-Fraction (Adaptive-$T$)}

Our second variant directly models the fraction of data retained at each epoch. Let $T_t \in (0,1]$ denote the proportion of the dataset used at epoch $t$, such that:
\[
|\mathcal{D}_t| = T_t \cdot N.
\]

We define a target decay trajectory $T_{\text{base}}(t)$ using a fixed schedule (e.g., logarithmic), but allow the actual trajectory $T_t$ to deviate from this schedule based on feedback.

At each epoch:
\begin{itemize}
    \item If $\Delta_t > \delta$, where $\delta$ is a small threshold, we follow the decay schedule:
    \[
    T_{t+1} = T_{\text{base}}(t+1).
    \]
    \item Otherwise, we apply a stochastic acceptance rule. If the update is rejected, we \textit{reheat} the system by increasing $T_t$, thereby increasing the dataset size:
    \[
    T_{t+1} = \min(T_0, \gamma \cdot T_t),
    \]
    where $\gamma > 1$ is a reheating factor.
\end{itemize}

This variant can be interpreted as tracking a desired curriculum while allowing deviations when learning stagnates. The reheating mechanism enables recovery from overly aggressive data reduction by temporarily increasing data exposure.

\subsection{Subset Sampling}

At each epoch, we construct $\mathcal{D}_t$ by randomly sampling $|\mathcal{D}_t|$ examples from $\mathcal{D}_0$ without replacement. Sampling is repeated independently across epochs, ensuring diversity while maintaining computational efficiency.

A visual depiction of the two variants can be seen in Figure~\ref{fig:overview}.

\begin{figure}
    \centering
    \includegraphics[width=0.99\linewidth]{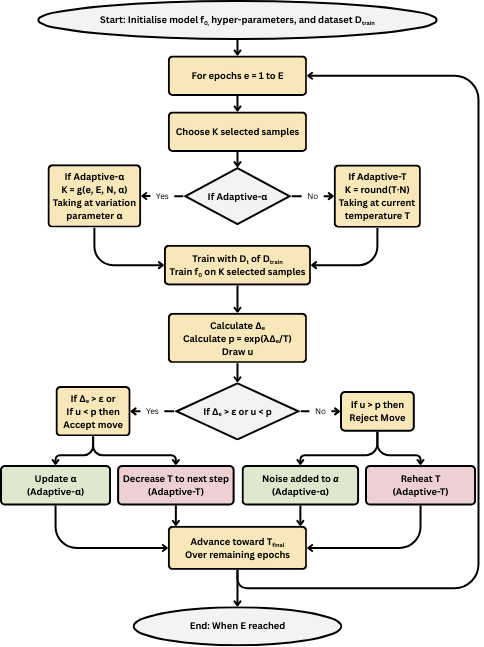}
    \caption{Adaptive data dropout dynamically selects training samples based on performance feedback, updating either the revision parameter ($\alpha$) or temperature ($T$) through an accept–reject mechanism to balance exploration and efficient learning.}
    \label{fig:overview}
\end{figure}

\subsection{Effective Epochs}

We adopt the notion of \textit{Effective Epochs} (EE) from~\cite{shriramprogressive} to quantify training efficiency. While a forward pass is performed on all samples, backpropagation is applied only to the selected subset. The total number of effective epochs is defined as:
\[
\text{Effective Epochs} = 
\frac{\sum_{t=1}^{T} |\mathcal{D}_t|}{N}.
\]

This metric reflects the equivalent number of full passes over the dataset and provides a hardware-independent measure of computational cost.

\subsection{Discussion}

Adaptive Data Dropout generalizes PDD by replacing fixed schedules with feedback-driven updates. From a cognitive perspective, this mirrors self-regulated learning, where learners adjust effort based on performance. By dynamically controlling data exposure, our method maintains an appropriate level of difficulty throughout training, balancing efficiency and generalization without requiring explicit difficulty estimation or precomputed schedules.

\subsection{Theoretical Insight}

Adaptive Data Dropout can be interpreted as a feedback-driven control process that dynamically regulates the effective training distribution. At each epoch, the model observes a performance signal $\Delta_t$ and adjusts the subset size $|D_t|$ accordingly. This induces a non-stationary training distribution that depends on the state of the model.

From an optimization perspective, this can be viewed as modulating the stochastic gradient noise. Reducing the dataset size increases gradient variance, encouraging exploration of the loss landscape, while increasing data exposure reduces variance and promotes stable convergence. The adaptive mechanism therefore alternates between exploration and consolidation phases, similar to simulated annealing, but driven by performance feedback rather than a fixed schedule.

The acceptance rule further introduces a stochastic decision process that allows occasional acceptance of non-improving updates. This prevents the model from becoming trapped in suboptimal regimes caused by overly aggressive data reduction. In the Adaptive-$T$ variant, the reheating mechanism explicitly increases data exposure when learning stagnates, enabling recovery and re-stabilization of training.

More broadly, the method can be interpreted as maintaining an implicit balance between bias and variance during optimization. When performance improves, the model tolerates higher bias (less data) to gain efficiency, whereas when performance deteriorates, it reduces bias by incorporating more data. This dynamic adjustment aligns with the principle of maintaining an appropriate level of learning difficulty, which has been shown to be critical in both human and machine learning systems.

While a full theoretical characterization remains an open problem, these perspectives suggest that adaptive data selection acts as a form of implicit regularization, shaping both the optimization trajectory and the generalization behavior of the model.

\begin{figure}
    \centering
    \includegraphics[width=0.99\linewidth]{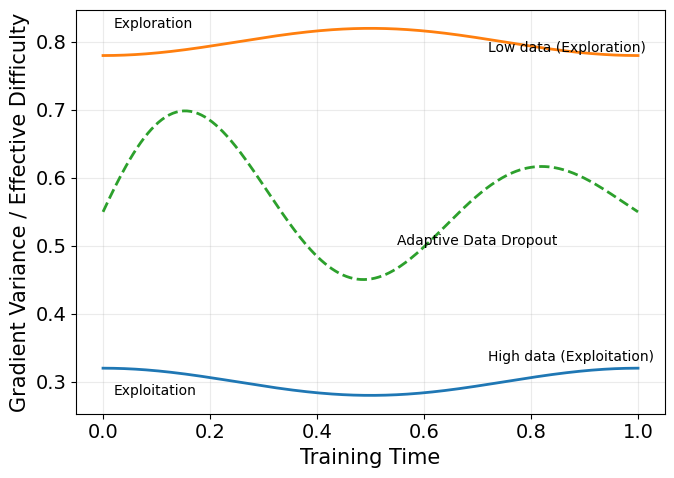}
    \caption{Conceptual illustration of adaptive data exposure as a balance between exploration and exploitation. Using more data leads to lower gradient variance and stable convergence, while using less data increases variance and promotes exploration. Adaptive Data Dropout dynamically transitions between these regimes based on training feedback, enabling efficient and robust optimization.}
    \label{fig:theory}
\end{figure}

Figure~\ref{fig:theory} illustrates this behavior conceptually. Unlike fixed schedules, which remain in either high- or low-variance regimes, Adaptive Data Dropout transitions between them over time. This behavior is analogous to annealing-based optimization, but driven directly by training feedback rather than a predefined schedule, enabling a more responsive and efficient learning process.

\section{Experimental Analysis}

\subsection{Datasets}

We evaluate our method on three standard image classification benchmarks: CIFAR-10~\cite{krizhevsky2009learning}, CIFAR-100~\cite{krizhevsky2009learning}, and ImageNet~\cite{deng2009imagenet}. The CIFAR-10 and CIFAR-100 datasets each contain $60{,}000$ color images of size $32 \times 32$, with $50{,}000$ training images and $10{,}000$ test images. CIFAR-10 consists of 10 object categories, while CIFAR-100 contains 100 finer-grained classes. These datasets are widely used for evaluating compact and mid-scale architectures. We also evaluate on ImageNet (ILSVRC2012), which contains over $1.28$ million training images and $50{,}000$ validation images spanning 1,000 categories. Compared to CIFAR, ImageNet provides substantially greater scale and diversity, making it a strong benchmark for assessing both generalization and training efficiency.

\subsection{Implementation Details}

Adaptive Data Dropout is implemented with minimal changes to standard training pipelines. We do not modify the underlying model architectures, loss functions, or optimizers, and only alter how subsets of training data are selected at each epoch. To preserve stability, we keep batch normalization behavior, learning-rate schedules, and augmentation policies consistent with the corresponding baseline training settings. For all variants with data reduction, the data loader is refreshed at each epoch to reflect the current sampled subset.

We evaluate on popular image classification architectures including MobileNetV2~\cite{sandler2018mobilenetv2}, EfficientNet~\cite{tan2019efficientnet}, EfficientFormer~\cite{li2022efficientformer}, ResNet~\cite{he2016deep}, and ViT-B~\cite{dosovitskiy2020image}. We follow the official implementation details for ImageNet experiments, and maintain the same optimization settings used in prior work for fair comparison. For CIFAR experiments, we use AdamW with a StepLR scheduler and standard augmentation settings. Our method is architecture-agnostic and integrates directly into existing supervised training pipelines.

We consider two adaptive variants in our experiments: \textbf{Adaptive-$\alpha$}, which dynamically adjusts the decay parameter of the subset schedule, and \textbf{Adaptive-$T$}, which directly controls the keep-fraction of the training data throughout training. In all cases, the final epoch uses the full dataset as a revision phase, consistent with the training principle introduced in Progressive Data Dropout (PDD)~\cite{shriramprogressive}.

\subsection{Supervised Image Classification}

In Table~\ref{tab:cifar}, we compare the performance of EfficientNet-B0~\cite{tan2019efficientnet}, MobileNet-V2~\cite{sandler2018mobilenetv2}, ResNet-50~\cite{he2016deep}, EfficientFormer-L1~\cite{li2022efficientformer}, and ViT-B-MAE~\cite{he2022masked} across CIFAR-10, CIFAR-100, ImageNet, and ImageNet-pretrained transfer settings. We report results for each model under the baseline training regime, using DBPD~\cite{shriramprogressive} as a representative static data dropout baseline as it provides a strong and interpretable baseline, as well as our two adaptive variants, Adaptive-$\alpha$ and Adaptive-$T$.

Our goal is to examine whether feedback-driven data scheduling provides a stronger trade-off between accuracy and computational cost than fixed schedules. Across datasets and architectures, the static data dropout methods already offer substantial savings over conventional training. We use these as strong baselines and evaluate whether adaptation can further improve robustness and efficiency by dynamically modulating data exposure in response to learning progress.

\begin{table*}
\caption{Accuracy (\%) and effective epochs reported as \textit{accuracy/effective epochs}. All methods besides ViT-MAE are trained from scratch unless otherwise mentioned. ViT-MAE uses the publicly available checkpoint~\cite{he2022masked}. All results have been conducted following official implementations on timm~\cite{timm}. Best results in \textcolor{blue}{blue} and second best in \textcolor{green}{green}. The last column shows the gains in accuracy and drop in effective epochs of Adaptive Data Dropout relative to the baseline.}
\resizebox{\linewidth}{!}{
\begin{tabular}{l ccccc cc}
\midrule
                       & Baseline & DBPD & Adaptive-$\alpha$ & Adaptive-$T$ & Acc. gain (\%) & EE saved \\ \midrule
\multicolumn{1}{l}{\textit{CIFAR10}}     &  &   &  &  &     &         \\ 
\multicolumn{1}{l}{EfficientNet-B0} & \textcolor{blue}{88.21} / 30 & \textcolor{green}{87.82} / \textcolor{blue}{5.86} & 85.69 / 7.45 & 85.28 / \textcolor{green}{6.36} & \textit{-2.52} & \textit{4.72} $\times$\\
\multicolumn{1}{l}{MobileNet-V2}   & \textcolor{blue}{85.49} / 30 & \textcolor{green}{85.39} / \textcolor{green}{6.42} & 83.08 / 8.76 & 80.79 / \textcolor{blue}{6.14} & \textit{-2.41}  & \textit{4.89} $\times$       \\
ResNet-50 & \textcolor{green}{84.85} / 30 & \textcolor{blue}{85.61} / \textcolor{green}{7.08} & 83.82 / 8.48 & 82.64 / \textcolor{blue}{6.24} & \textit{-2.21} & \textit{4.81} $\times$ \\ 
EfficientFormer-L1 & \textcolor{green}{80.32} / 30 & \textcolor{blue}{81.17} / \textcolor{blue}{5.82} & 80.75 / 7.21 & 80.67 / \textcolor{green}{6.90} & \textit{0.43} &  \textit{5.15} $\times$ \\ \midrule

\multicolumn{1}{l}{\textit{CIFAR100}}     &  &   &  &  &     &     \\ 
\multicolumn{1}{l}{EfficientNet-B0} & \textcolor{green}{66.32} / 200 & \textcolor{blue}{67.15} / \textcolor{green}{24.8}& 66.51 / 23.94 & 63.67 / \textcolor{blue}{21.13} & \textit{0.19} & \textit{8.36} $\times$ \\
\multicolumn{1}{l}{MobileNet-V2} &    \textcolor{green}{61.25} / 200 & 62.85 / \textcolor{green}{29.6} & \textcolor{blue}{63.14} / 25.61 & 62.99 / \textcolor{blue}{22.62} & \textit{1.89} & \textit{7.84} $\times$\\
ResNet-50       & \textcolor{green}{59.81} / 200 & \textcolor{blue}{60.13} / 16.26 & 59.67 / 16.11 & 59.41 / \textcolor{blue}{13.78} & \textit{-0.14} & \textit{14.51} $\times$ \\ 
EfficientFormer-L1  & 55.91 / 200 & \textcolor{green}{57.62} / \textcolor{green}{30} & \textcolor{blue}{58.11} / 31.31 & 55.95 / \textcolor{blue}{24.67} & \textit{2.20} & \textit{8.11} $\times$ \\
\midrule

\multicolumn{3}{l}{\textit{CIFAR100 finetuned from ImageNet}}     &  &  &    &         \\ 
\multicolumn{1}{l}{EfficientNet-B0} & \textcolor{green}{83.31} / 200 & 83.95 / \textcolor{green}{25.4} & \textcolor{green}{83.98} / 26.1 & \textcolor{blue}{84.11} / \textcolor{blue}{20.8} & \textit{0.80} & \textit{9.62} $\times$\\
\multicolumn{1}{l}{MobileNet-V2}    & 74.10 / 200 & \textcolor{green}{74.35} / \textcolor{blue}{22.4} & \textcolor{blue}{74.61} / 25.3  & 74.39 / \textcolor{green}{19.5} & \textit{0.51} & \textit{10.26} $\times$ \\
ResNet-50 & 78.59 / 200 & \textcolor{green}{79.92} / \textcolor{blue}{21} & \textcolor{blue}{80.17} / 23.4 & 79.97 / \textcolor{green}{17.8} & \textit{1.58} & \textit{11.24} $\times$ \\ 
EfficientFormer-L1 & 85.30 / 200 & \textcolor{green}{86.74} / \textcolor{green}{36.6} & \textcolor{blue}{87.28} / 39.8 & 86.85 / \textcolor{blue}{33.2} & \textit{1.98} & \textit{6.02} $\times$\\
ViT-B-MAE  & 86.81 / 200 & 85.92 / \textcolor{green}{52.2} & \textcolor{blue}{86.94} / 55.8 & 85.55 / \textcolor{blue}{43.5} & \textit{0.13} & \textit{4.60} $\times$\\
\midrule

\multicolumn{1}{l}{\textit{ImageNet}}     &  &   &  &  &     &      \\ 
EfficientNet-B0 & 77.10 / 350 & \textcolor{green}{77.45} / \textcolor{green}{111.2} & \textcolor{blue}{78.15} / 134.2 & 77.71 / \textcolor{blue}{105.5}  & \textit{1.05} & \textit{3.32} $\times$ \\
MobileNet-V2 & \textcolor{blue}{71.60} / 250 & 68.42 / \textcolor{green}{54.5} & \textcolor{green}{70.82} / 53.9 & 67.59 / \textcolor{blue}{45.8} & \textit{-0.78} & \textit{5.46} $\times$ \\
ResNet-50 & 75.04 / 100 & \textcolor{green}{76.21} / \textcolor{green}{33.0} & \textcolor{blue}{76.82} / 35.5 & 76.25 / \textcolor{blue}{31.5} & \textit{1.78} & \textit{3.17} $\times$ \\ 
EfficientFormer-L1 & 79.11 / 300 & \textcolor{green}{79.24} / \textcolor{green}{91.5} & \textcolor{blue}{79.45} / 95.8 & 78.87 / \textcolor{blue}{85.7} & \textit{0.34} & \textit{3.50} $\times$ \\
ViT-B-MAE & 83.10 / 100 & \textcolor{green}{83.31} / \textcolor{green}{35.4} & \textcolor{blue}{83.81} / 37.8  & 83.06 / \textcolor{blue}{27.5} & \textit{0.71} & \textit{3.64} $\times$ \\
\bottomrule
\end{tabular}}
\label{tab:cifar}
\end{table*}

\begin{figure}
    \centering
    \includegraphics[width=0.99\linewidth]{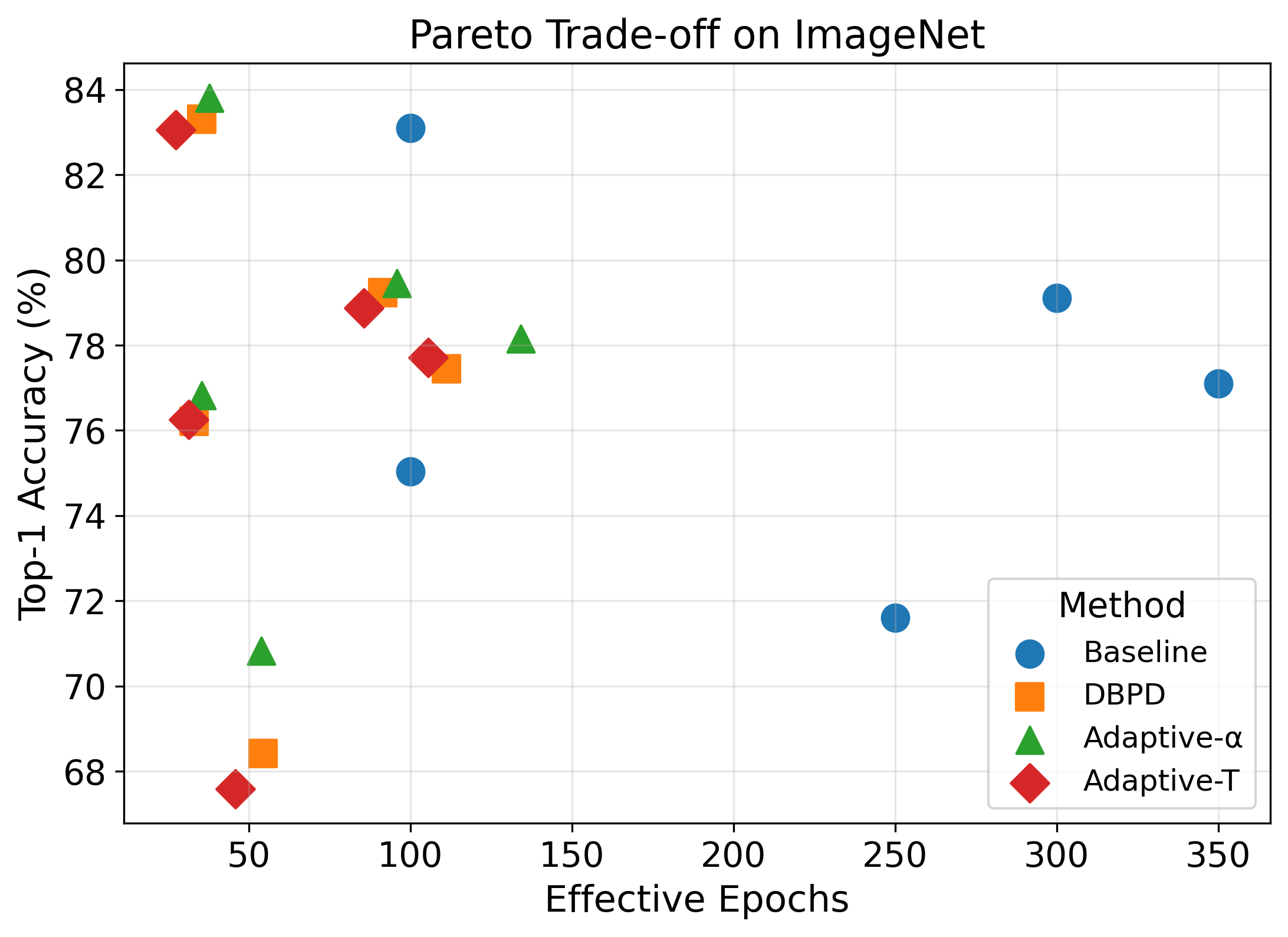}
    \caption{Pareto trade-off between Top-1 accuracy and effective epochs on ImageNet. Each point represents a model, with colors indicating architectures and markers denoting methods. Adaptive Data Dropout consistently achieves a better accuracy–efficiency trade-off, achieving competitive accuracy while using substantially fewer effective training epochs.}
    \label{fig:pareto}
\end{figure}

To further illustrate this trade-off, we visualize the relationship between accuracy and effective training cost in Figure~\ref{fig:pareto}. The Pareto plot highlights that Adaptive Data Dropout consistently achieves competitive or improved accuracy while requiring substantially fewer effective epochs compared to both standard training and fixed data dropout schedules. Notably, the adaptive variants occupy more favorable regions of the Pareto frontier, demonstrating that feedback-driven data selection enables a more efficient use of training data without sacrificing performance. These results reinforce the benefit of dynamically adjusting data exposure in response to training progress.

\subsection{Comparison to State-of-the-Art}

We compare Adaptive Data Dropout against strong baselines for efficient training and data selection, including DataDiet (ELN and Forget)~\cite{paul2021deep}, IES~\cite{yuan2025instancedependent}, Early Stopping, and InfoBatch~\cite{qininfobatch}. We follow the official implementations for all baselines and use effective epochs to evaluate efficiency. This comparison is designed to assess whether adaptive control of data exposure provides benefits beyond fixed subset schedules, heuristic importance estimation, or early termination strategies. The results are reported on CIFAR-100 fine-tuning from ImageNet under a common training setup.

\begin{table*}[h]
\centering
\caption{Comparison with strong active learning baselines on CIFAR-100. Best results are shown in \textcolor{blue}{blue} and second best in \textcolor{green}{green}. DDE corresponds to DataDiet ELN and DDF corresponds to DataDiet Forget.}
\label{tab:active_learning_cifar100}
\resizebox{\linewidth}{!}{
\begin{tabular}{l|c|c|c|c|c|c|c|c|c}
\toprule
\textbf{Model} & \textbf{Baseline} & \textbf{DDE} & \textbf{DDF} & \textbf{IES} & \textbf{Early Stopping} & \textbf{InfoBatch} & \textbf{DBPD} & \textbf{Adaptive-$\alpha$} & \textbf{Adaptive-$T$} \\
\midrule
ResNet-50 
& 78.5 / 200  
& 78.6 / 141  
& 79.1 / 142  
& 79.5 / 157  
& 74.3 / 121  
& \textcolor{blue}{80.4} / 145  
& \textcolor{green}{80.0} / \textcolor{green}{21} 
& 80.2 / 23.4 
& 80.0 / \textcolor{blue}{17.8} \\

EfficientNet 
& 83.3 / 200  
& 82.7 / 139  
& 83.0 / 139  
& 83.7 / 153  
& 76.6 / 132  
& \textcolor{green}{84.0} / 141  
& \textcolor{green}{84.0} / \textcolor{green}{25.1} 
& \textcolor{green}{84.0} / 26.1 
& \textcolor{blue}{84.1} / \textcolor{blue}{20.8} \\

EfficientFormer-L1 
& 85.3 / 200  
& 84.8 / 142  
& 85.0 / 140  
& 85.8 / 155  
& 78.2 / 125  
& 86.6 / 143  
& 86.8 / \textcolor{green}{36.6} 
& \textcolor{blue}{87.3} / 39.8 
& \textcolor{green}{86.9} / \textcolor{blue}{33.2} \\

\bottomrule
\end{tabular}}
\end{table*}

\vspace{-0.2cm}
\subsection{Ablation Study}

To better understand the behavior of Adaptive Data Dropout, we conduct an ablation study along several dimensions, including matched-step baseline comparisons, sensitivity to adaptive control parameters, shortened training schedules, and the effect of revision. Unless otherwise stated, all ablations are performed using EfficientNet-B0 trained from scratch on CIFAR-100.

\textbf{Are the gains simply due to fewer effective epochs?}  
To ensure that any improvement is not merely a consequence of reduced optimization steps, we train a baseline EfficientNet-B0 for the same number of effective epochs as each dropout variant. This comparison isolates whether performance gains arise from the schedule itself rather than from training less. We report these results in Table~\ref{tab:matched_baseline}.

\begin{table}[t]
\centering
\caption{Baseline trained to match the effective number of epochs (EE) used by dropout variants.}
\resizebox{\linewidth}{!}{
\begin{tabular}{l|c|c|c}
\toprule
\textbf{Variant} & \textbf{EE} & \textbf{Baseline (\%)} & \textbf{Accuracy (\%)} \\
\midrule
Adaptive-$\alpha$ & 23.9 & 63.18 & 66.51 \\
Adaptive-$T$ & 21.1 & 61.82 & 63.67 \\
\bottomrule
\end{tabular}
}
\label{tab:matched_baseline}
\end{table}

\textbf{Can we shorten training and still retain performance?}  
To assess performance under constrained training budgets, we repeat all dropout variants under shorter training schedules and compare them to baselines trained for the same number of epochs. This experiment evaluates whether adaptive data exposure remains useful when optimization time is limited.

\begin{table*}[t]
\centering
\caption{Accuracy (\%) and effective epochs reported as \textit{accuracy/effective epochs} for CIFAR-100 with EfficientNet-B0, trained for 25 and 50 epochs from scratch.}
\begin{tabular}{l|c|c|c|c}
\toprule
\textbf{Model } & \textbf{Baseline} & \textbf{DBPD 0.7} & \textbf{Adaptive-$\alpha$} & \textbf{Adaptive-$T$} \\
\midrule
(50 epochs) &&&& \\
EfficientNet-B0 (C100) & 63.65 / 50 & 67.62 / 18.64 & 67.95 / 20.71 & 66.58 / 14.38 \\
\hline
(25 epochs)  &&&& \\
EfficientNet-B0 (C100) & 63.32 / 25 & 66.94 / 14.02 & 66.85 / 15.24 & 66.07 / 10.51 \\
\bottomrule
\end{tabular}
\label{tab:ablation_50epoch}
\end{table*}

\subsection{Dynamics of Adaptive Reheating} 

To better understand the behavior of Adaptive Data Dropout, we analyze how the number of training samples evolves over epochs. In particular, we examine whether the proposed feedback-driven mechanism exhibits the intended ``reheating'' behavior, where data exposure is increased when learning stagnates or deteriorates.

\begin{figure*}[t]
\centering
\begin{minipage}{0.48\linewidth}
    \centering
    \includegraphics[width=\linewidth]{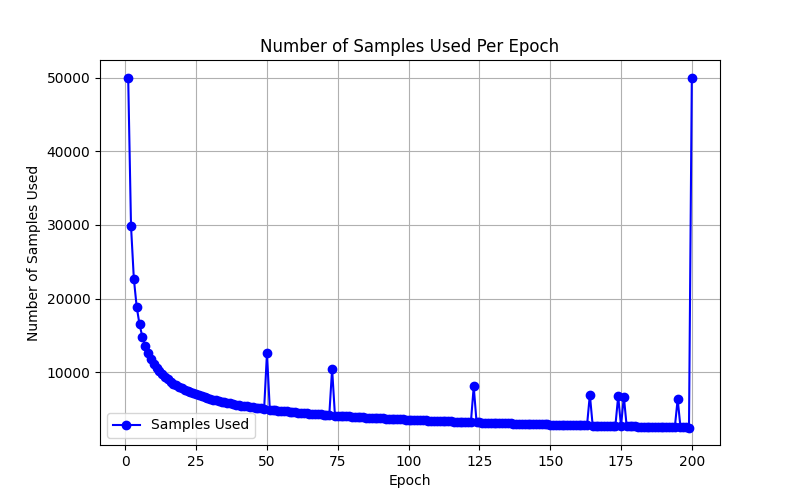}
    \vspace{-2mm}
    \caption*{(a) EfficientNet-B0}
\end{minipage}
\hfill
\begin{minipage}{0.48\linewidth}
    \centering
    \includegraphics[width=\linewidth]{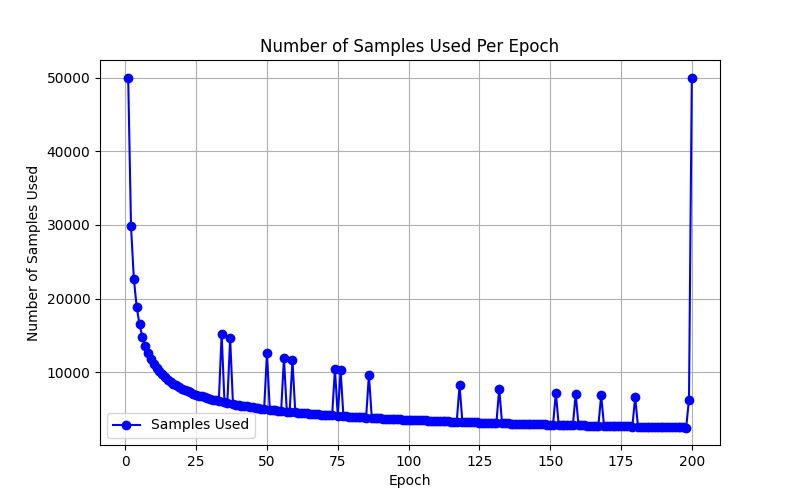}
    \vspace{-2mm}
    \caption*{(b) ResNet-50}
\end{minipage}

\caption{Number of training samples used per epoch under Adaptive Data Dropout. The model progressively reduces data usage but intermittently increases it in response to stagnation or drops in performance, demonstrating adaptive reheating behavior.}
\label{fig:reheating}
\vspace{-4mm}
\end{figure*}

Figure~\ref{fig:reheating} visualizes the number of samples used per epoch for two representative models. We observe a consistent pattern across both variants: the training process initially reduces the dataset size as performance improves, reflecting increasing confidence and redundancy in the data. However, at several points during training, sharp increases in the number of samples occur. These spikes correspond to periods where the improvement in training accuracy slows down or reverses, triggering the adaptive mechanism to increase data exposure.

This behavior closely mirrors the notion of self-regulated learning, where learners allocate more effort when performance deteriorates and reduce effort when learning is progressing well. Importantly, these reheating events are not pre-scheduled but emerge naturally from the feedback signal, demonstrating that the model dynamically adjusts its learning difficulty over time.

Overall, these results provide empirical evidence that Adaptive Data Dropout does not follow a fixed curriculum, but instead continuously adapts data usage in response to training dynamics. This ability to recover from overly aggressive data reduction is a key advantage over static schedules such as Progressive Data Dropout.

\section{Limitations and Future Work}
\label{sec:limitations}

Although Adaptive Data Dropout addresses an important limitation of fixed data schedules, it still has several constraints. First, the method introduces additional control hyperparameters, such as update thresholds, decay functions, and reheating factors, which may require careful tuning across different datasets, architectures, and optimization settings. While these parameters are relatively lightweight compared to model-level modifications, their interaction with training dynamics can influence both stability and efficiency. Second, because the adaptation mechanism is driven by training feedback, its effectiveness depends on the quality and stability of the chosen performance signal. In particular, noisy or highly fluctuating accuracy estimates, especially in early training, may lead to suboptimal or unstable updates. Incorporating more robust or aggregated signals, such as smoothed accuracy, loss trends, or uncertainty measures, may improve reliability.

Third, although our experiments demonstrate consistent gains across standard image classification benchmarks, broader validation on other vision tasks, such as detection, segmentation, and video understanding, is necessary to fully establish the generality of the framework. These settings may exhibit different training dynamics, where the role of adaptive data exposure could differ.

Finally, while the adaptive strategy improves flexibility compared to fixed schedules, it still requires selecting an initial schedule family or operating range, which implicitly constrains the space of possible behaviors. Future work could explore more principled and autonomous controllers that reduce or eliminate this dependency. In particular, one promising direction is to learn the adaptation policy itself, for example through reinforcement learning or meta-learning, enabling the model to discover optimal data scheduling strategies directly from training signals. Such approaches could further unify data selection with optimization, potentially leading to fully self-regulating training systems that adapt not only data exposure but also other aspects of the learning process, such as augmentation, sampling, or learning rate schedules.

\section{Conclusion}

We introduced \textit{Adaptive Data Dropout}, a simple and effective extension of Progressive Data Dropout that replaces fixed schedules with feedback-driven adaptation. Inspired by self-regulated learning, our method dynamically adjusts data exposure during training based on observed progress, enabling the model to regulate the difficulty of learning over time. We considered two adaptive variants, Adaptive-$\alpha$ and Adaptive-$T$, which respectively control the decay parameter and the keep-fraction of the training data. Across standard image classification benchmarks and architectures, our framework is designed to provide a more favorable trade-off between accuracy and computational efficiency relative to static data dropout strategies. These results support the broader view that training dynamics can benefit from adaptive, human-inspired principles rather than fixed curricula alone.
{
    \small
    \bibliographystyle{ieeenat_fullname}
    \bibliography{main}
}


\end{document}